\documentclass[runningheads]{llncs}
\usepackage[T1]{fontenc}
\usepackage{amsmath}
\usepackage{graphicx}
\usepackage[table,xcdraw]{xcolor}
\usepackage[font=small,skip=2pt]{caption}
\usepackage[skip=0pt]{subcaption}
\usepackage{multicol}
\usepackage{multirow}
\usepackage{adjustbox}
\usepackage{hyperref}
\usepackage{color}

\begin{document}
\title{Mitigating Vanishing Activations in Deep CapsNets Using Channel Pruning}
\titlerunning{Mitigating Vanishing Activations in CapsNet Using Pruning}
\author{Siddharth Sahu\orcidID{0009-0005-1835-0492}
\and
Abdulrahman Altahhan\orcidID{0000-0003-1133-7744}}
\authorrunning{S. Sahu and A. Altahhan}
\institute{University of Leeds, Leeds, LS2 9JT United Kingdom \\
\email{\{od21ss, a.altahhan\}@leeds.ac.uk}}
\bibliographystyle{splncs04}

\maketitle
\begin{abstract}
Capsule Networks outperform Convolutional Neural Networks in learning the part-whole relationships with viewpoint invariance, and the credit goes to their multidimensional capsules. It was assumed that increasing the number of capsule layers in the capsule networks would enhance the model performance. However, recent studies found that Capsule Networks lack scalability due to vanishing activations in the capsules of deeper layers. This paper thoroughly investigates the vanishing activation problem in deep Capsule Networks. To analyze this issue and understand how increasing capsule dimensions can facilitate deeper networks, various Capsule Network models are constructed and evaluated with different numbers of capsules, capsule dimensions, and intermediate layers for this paper. Unlike traditional model pruning, which reduces the number of model parameters and expedites model training, this study uses pruning to mitigate the vanishing activations in the deeper capsule layers. In addition, the backbone network and capsule layers are pruned with different pruning ratios to reduce the number of inactive capsules and achieve better model accuracy than the unpruned models.

\keywords{Capsule Network \and Dynamic Routing \and Channel Pruning}
\end{abstract}

\section{Introduction}
The human brain learns part-whole relationships quite well with viewpoint invariance while recognizing shapes. Inspired by this philosophy, Hinton \textit{et al.} \cite{Hinton2011} proposed the concept of ‘capsules’ that can recognize implicit entities by computing intrinsic visual information in multiple dimensions of the appearance manifold. The proposal was novel to the artificial neural network community because the capsules generate vector outputs that contain diverse spatial information instead of the scalar output produced by neurons of the artificial neural networks. %In those days, networks used for object detection and pattern recognition tasks were excellent at object detection tasks that did not need the knowledge of precise spatial relationships between the high-level parts. Thus, the advent of the concept of capsules was highly promising.

Convolutional neural networks (CNNs) are widely used to recognize objects and patterns in images. They are computationally efficient and show strong translational equivariance due to convolution operations and sparse weight sharing. However, CNNs fail to generalize their knowledge to novel viewpoints, contributing to poor performance in shape recognition and overlapping object segmentation tasks \cite{Hinton2011}. CNNs' inability to handle viewpoint variations makes them highly susceptible to adversarial attacks. Many researchers have worked on achieving viewpoint invariance and equivariance for CNNs. For example, \cite{Gens2014} built deep symmetry networks, \cite{Lenc2015} proposed DetNet to detect transformations from different viewpoints, and \cite{Worrall2017} introduced Harmonic Networks that used circular harmonic filters. Most of the efforts are to add rotational invariance using data augmentation, but a few have worked on improving the general viewpoint equivariance.

Several years after the "capsules" were introduced \cite{Hinton2011}, \cite{Sabour2017} proposed an architecture of the Capsule Neural Network (CapsNet) that outperformed CNNs in overlapping digit segmentation. The proposed network used an iterative routing-by-agreement method called Dynamic Routing (DR) to pass the output of lower-level capsules to higher-level effectively. 
Using DR, CapsNet preserves more information from lower-level capsules than CNNs' max-pooling, which passes only the most prominent features from the lower layers to the upper layers.
The capsule output is a vector representing the input data's contribution to different object properties, like position, size, orientation, texture, etc., which enables dynamic routing of the individual capsule outputs in a lower layer to appropriate higher-layer capsules. However, when more capsule layers were added to CapsNet, it was observed that the deep CapsNet suffered from vanishing activations, resulting in low network utilization and poor scalability \cite{Mitterreiter2023}. 

To mitigate vanishing activations, many routing algorithms have been proposed for the Capsule Networks to date. In 2018, \cite{Hinton2018} introduced a routing based on the Expectation-Maximization (EM) algorithm, which iteratively updates the votes of the lower capsules using the means, variances, and activation probabilities of the capsules in the layer above. EM routing makes the Capsule Networks more robust to white-box adversarial attacks than a baseline CNN.  The Self-Routing (SR) algorithm eliminated the concept of routing-by-agreement between capsules in the same layer and proposed an independent capsule routing network consisting of a single-layer perceptron to generate individual capsule routing coefficients \cite{Hahn2019}. SR outperforms dynamic \cite{Sabour2017} and EM \cite{Hinton2018} routing even when the model size increases, and made the CapsNet the most robust for adversarial attacks. After that, \cite{Ribeiro2020} presented an idea of Variational Bayes (VB) routing that uses a mixture of transforming Gaussians to learn and model the uncertainties of the capsule parameters and routing coefficients, reducing the variance-collapse issues in EM routing.

\cite{Wang2018} demonstrated that CapsNet dynamic routing can be solved as an optimization problem, and better performance can be achieved if KL divergence is used to pass the output of lower-level capsules to higher-level capsules effectively. SegCaps \cite{lalonde2018capsules} used 2D convolutions for voting. \cite{Rajasegaran2019} introduced a deep Capsule Network called DeepCaps which implements dynamic routing with skip connections for reducing vanishing gradients and 3D convolutions for high-quality votes generation. DeepCaps performs comparatively well on complex datasets like CIFAR10 albeit the increase in performance is not significant. 

Last but not least, \cite{Everett2023} carried out an exhaustive study of the performance of all the different capsule routing algorithms and their effects on the vanishing activations and found that irrespective of the routing algorithm used, deep Capsule Networks suffer from scalability issues due to the vanishing activations problem. 
%The authors doubted whether routing is the real culprit behind the scalability issues or if other inherent problems in the network architecture exist. Thinking along the same lines, 
While pevious research has primarily focused on alleviating the vanishing activations in Capsule Networks by improving the routing algorithm, this paper takes a different approach. We investigate pruning convolutional channels using their importance evaluation \cite{Sui2021} along with training the CapsNet with Correlation Coefficient Matrix (CCM) loss \cite{Wang2023}. The combined strategy aims to address the vanishing activations problem in deep CapsNet. 

\section{Background}
Pruning is widely applied to CNN models to reduce computation cost and memory usage without significantly compromising the model's accuracy. Pruning can be performed in two different ways: structured and unstructured. On the one hand, structured pruning ensures that the network maintains its topology post-pruning and usually involves removing the filters or entire convolutional layers. \cite{Li2017} introduced filter pruning in CNNs and showed that the training accuracies of models like VGG-16 and ResNet-56 are not much affected even after pruning 34. 2\% and 27. 6\% of their tuning parameters, respectively. 

On the other hand, unstructured pruning minimizes the model parameters by removing the non-contributing weights without paying any importance to the topology of the pruned network. The weight pruning performed on AlexNet and VGG-16 models showed impressive compression ratios of 35\% and 49\%, respectively, without impacting model accuracies \cite{Han2016}. Nevertheless, it may not reduce the computational complexity of convolutional layers due to the introduction of irregular sparsity after pruning \cite{Li2017}. \cite{Bragagnolo2021} identified that structured pruning 
%removes significantly fewer model parameters but 
results in lower model memory requirements and faster inference time than unstructured pruning. For this reason, structured pruning is adopted as the method of choice in this paper to prune Capsule Networks.

To improve the structured filter pruning performance, \cite{Wang2023} presented a data-driven CCM loss calculated using the correlation coefficient matrix between different feature maps in a network layer. 
%CCM loss differs from the losses calculated using L1 and L2 norms that focus on generating more zeros for increasing redundancy. 
CCM loss ensures the model learns stronger linear relations between the feature maps during training and compresses shared information into fewer useful channels for effective filter pruning. The authors also used the CHannel Independence-based Pruning (CHIP) technique, introduced by  \cite{Sui2021}, to prune the less important channels in CNN models. CHIP incorporates inter-channel information to prune the convolutional neural networks efficiently. It identifies channel importance using nuclear norms on their feature maps and selects channels that have feature maps with less information for pruning. The authors’ experiments show that CHIP produces high pruning performance while preserving or slightly increasing the model accuracy. Thus, this paper uses CHIP with CCM loss to prune the backbone network and the primary capsule layer during model training.

\cite{Renzulli2022} built the CapsNet using pruned backbone networks to reduce computational complexity and demonstrated that using structurally sparse backbone networks can improve the efficiency, memory consumption and training time of the overall Capsule Network. Moreover, they observed that occasionally adding more capsule layers increases accuracy. 
% However, there are two major differences between Renzulli and Grangetto \cite{Renzulli2022} research and this paper. Firstly, they focused on reducing the computational complexity of the overall Capsule Network using feature-map pruning. 
In contrast to \cite{Renzulli2022}, this paper addresses the challenge of vanishing activations in deeper CapsNet by implementing feature-map pruning of both the backbone network and the primary capsule layer. Furthermore, unlike the pruning approach taken by \cite{Renzulli2022}, the model in this work is pruned using CCM loss \cite{Wang2023} and the CHIP algorithm \cite{Sui2021}.

\section{Design and Methodology}
A simple CapsNet \cite{Sabour2017} is a shallow network consisting of three layers: a convolutional layer, a primary capsule layer, and a digit capsule layer. The first is a 2D convolutional layer to extract the feature maps from an input image. The second is a primary capsule layer (PrimaryCaps), which is also a convolutional layer, and the output of this layer contains n-dimensional vectors for all the individual capsules in any channel, unlike the scalar-valued feature maps generated by the 2D convolutional layer. The final layer is called the digit capsule layer (DigitCaps), as it contains as many capsules as the number of digit classes. The DigitCaps has a transformation matrix trained to map the output of all the capsules in the PrimaryCaps to the digit class capsules and uses a dynamic routing algorithm to produce the final vector output for each digit class capsule.

\subsection{Model Architecture}
Our model is similar to the CapsNet model proposed by \cite{Sabour2017} with two main differences. First, we add more convolutional layers to the backbone to increase its feature extraction ability and use a different loss to facilitate structured filter pruning performance similar to \cite{Wang2023}. Since the backbone network is responsible for detecting local features in the input image and passing it to the PrimaryCaps layer through the generated feature maps, we wanted to give our network more capabilities to detect these local features. Therefore, instead of using a 2D convolutional layer for the backbone network, we used three convolutional layers. Each layer has 3$\times$3 convolution kernels, a padding of 1, ReLU activation and batch normalization. Furthermore, we used a stride of 1 for the first and third convolutional layers and 2 for the second convolutional layer.
\begin{figure}
    \centering
    \includegraphics[width=0.65\textwidth, height=3.8cm]{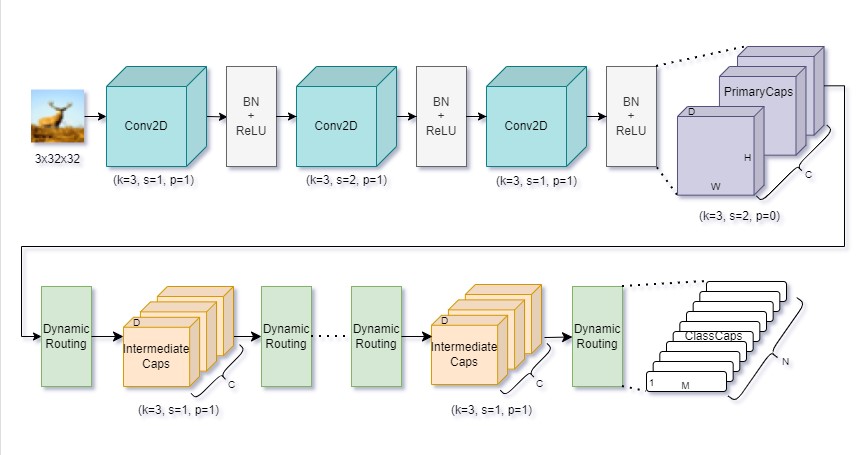}
    \caption{\label{fig:1}Capsule Network Architecture }
\end{figure}

The PrimaryCaps layer is another convolutional layer with a kernel of 3$\times$3, stride of 2, zero padding, and squashing activation. It has C$\times$D output channels, where C is the number of capsules, and D is the capsule's dimension in the layer. The layer generates the convolutional output of width and height, denoted by W and H, respectively, as shown in Fig.~\ref{fig:1}. Thus, the layer produces the W$\times$H$\times$C output capsules of D-dimension. The ClassCaps layer is the last layer of the CapsNet architecture used in this paper and is similar to the DigitCaps \cite{Sabour2017}. Its output is an N$\times$M Matrix, where N is the number of class capsules, and M is the dimension of the vector output of each class capsule.

To reproduce the vanishing activation problem \cite{Everett2023}, the depth of the Capsule Network must be increased by adding multiple intermediate capsule layers, also called IntermediateCaps, between the PrimaryCaps and the ClassCaps layers. Our implementation of IntermediateCaps layers follows the ConvCaps layer \cite{Hinton2018} used for the EM routing and has a transformation matrix like a ClassCaps layer. For IntermediateCaps, the number of capsules C, capsule output's width and height (W and H), and capsule dimension D are the same as those of the PrimaryCaps layer. The kernel's size, stride and padding for all the convolutional capsules in the IntermediateCaps layers are 3$\times$3, 1 and 1, respectively. As the number of IntermediateCaps layers increases, the computational complexity drastically increases. Thus, a maximum of 10 IntermediateCaps layers are added to the Capsule Network for this paper.

To learn the part-whole relationships in the given images, dynamic routing \cite{Sabour2017} between various capsule layers is chosen. %Other routing algorithms like EM \cite{Hinton2018}, VB \cite{Ribeiro2020} and SR \cite{Hahn2019} are not explored because. 
DR, like other routing algorithms such as  EM \cite{Hinton2018}, VB \cite{Ribeiro2020} and SR \cite{Hahn2019}, suffers from vanishing activations and it further provides us with a simple and robust platform for analyzing the effects of pruning on the vanishing activations of capsules.

\subsection{Dynamic Routing}
The dynamic routing algorithm is at the core of our CapsNet, which routes the outputs of the lower-layer capsules to the appropriate capsules in the layer above using an iterative process as presented in equation (\ref{eq:1}), where \(b_{ij}^{(r)}\) is the logits calculated between capsules \textit{i} in the lower layer and capsules \textit{j} in the layer above for the  \(r^{th}\) iteration.
\begin{equation} \label{eq:1}
    b_{ij}^{\left(r+1\right)}=b_{ij}^{\left(r\right)}+{\hat{u}}_{\left.j\right|i}\cdot squash\left(s_j\right)
\end{equation}
\begin{equation} \label{eq:2}
{\hat{u}}_{\left.j\right|i}=W_{ij}u_i
\end{equation}
The logits are initialized to zeros and iteratively updated by adding the product of \({\hat{u}}_{j|i}\) and squashed \(s_j\). 
\({\hat{u}}_{j|i}\) is obtained by multiplying capsule output \(u_i\) from the lower layer with the transformation matrix \(W_{ij}\) of the layer above, shown in equation (\ref{eq:2}). 
The transformed output \({\hat{u}}_{j|i}\), which holds the capsule predictions from the layer below, is then multiplied with the coupling coefficients \(c_{ij}\) calculated between the capsules of the two layers involved in the routing. The coupling coefficients \(c_{ij}\) are derived by applying the Softmax to the logits \( b_{ij} \) and are iteratively adjusted throughout the routing process when the logits are updated. For all the capsules in the lower layer, the weighted sum of the coupling coefficients and the transformed outputs are used to produce \(s_j\) as shown in equation (\ref{eq:3}). 
\begin{equation} \label{eq:3}
    s_j=\sum_{i}{c_{ij}{\hat{u}}_{\left.j\right|i}}
\end{equation}
\begin{equation} \label{eq:4}
    squash\left(s_j\right)=\frac{\|s_j\|_2}{1 + \|s_j\|_2} \cdot \frac{s_j}{\|s_j\|}
\end{equation}
To update \(b_{ij}\) according to the equation (\ref{eq:1}), \(s_j\) is squashed before multiplying it with \({\hat{u}}_{j|i}\) to introduce non-linearity in the capsule predictions as illustrated in equation (\ref{eq:4}). In the squashing operation, the output vectors of the capsules are squashed to shrink the length of the short vectors close to zero and the long vectors close to one \cite{Sabour2017}. The final output vectors obtained after several iterations are considered the output of dynamic routing.

\subsection{Loss Function}
This paper uses a combination of two loss functions for model training: the capsule margin loss \cite{Sabour2017} and the CCM loss \cite{Wang2023}. The total loss function is calculated by subtracting a small portion of the total CCM loss from the total capsule margin loss, as shown in equation (\ref{eq:5}), where the first loss term is for the total margin loss and the second is for the CCM loss.
\begin{equation} \label{eq:5}
    {Loss}_{total}=\sum_{n}^{N}{Loss}_{margin\ n}-\alpha\sum_{l}^{L}{Loss}_{ccm}^{\left(l\right)}
\end{equation}
The total margin loss of the model is obtained by adding the individual margin losses of all the N class capsules in the ClassCaps layer. For each class capsule $n$, a separate margin loss is calculated using the equation (\ref{eq:6}), where \(\|v_n\|\) is the norm of the M-dimensional output vector of the \(n^{th}\) class capsule, \(T_n\) is 1 when the class $n$ is present in the capsule output and 0 otherwise,  \(t_{pos}\) = 0.9,  \(t_{neg}\) = 0.1, and \(\lambda\) = 0.5.
\begin{equation} \label{eq:6}
    {Loss}_{margin\ n}=T_nmax(0,t_{pos}-\|v_n\|)^2+\lambda(1-T_n)max(0,\|v_n\|-t_{neg})^2
\end{equation}
\begin{equation} \label{eq:7}
    {Loss}_{ccm}^{\left(l\right)}=\frac{1}{C^{\left(l\right)}\times\ C^{\left(l\right)}}\sum_{i}^{C^{\left(l\right)}}\sum_{j}^{C^{\left(l\right)}}\left|corr\left(f_i^{\left(l\right)},f_j^{\left(l\right)}\right)\right|
\end{equation}
The total CCM loss is calculated by adding the CCM losses of all the convolutional layers in the backbone network and the PrimaryCaps. This loss enables the convolutional layer to learn linear relationships between the feature maps during the training \cite{Wang2023}. The formula for calculating CCM loss for any convolutional layer \textit{l} is given in equation (\ref{eq:7}), where \(C^{(l)}\) is the number of channels in the convolutional layer \textit{l}, and \(corr\left(f_i^{\left(l\right)},f_j^{\left(l\right)}\right)\) is the correlation coefficient calculated for the feature maps \textit{i} and \textit{j} of the convolutional layer \textit{l}.

The layer-wise CCM loss is calculated by averaging the correlation coefficient matrices for all the layer's feature map pairs. The total CCM loss is multiplied by a small constant \(\alpha\). The constant \(\alpha\) decides the weightage of the CCM loss in the overall loss calculation carried out during model training.

\subsection{Pruning}
In order to mitigate the vanishing activations observed in the deep CapsNet, the convolutional \textit{channels} in the backbone network and the PrimaryCaps layer are pruned. Moreover, for the effective pruning operation, the model is trained with CCM loss before pruning, which is essential to ensure that enough linear redundancy exists among the channels of a convolutional layer before their channel independence (CI) scores are calculated using the CHIP algorithm.

A channel CI score reflects the effect of omitting that channel feature map from the set of all feature maps of a layer. According to the CHIP algorithm \cite{Sui2021}, the nuclear norm of all channels in a convolutional layer is calculated first. Then, one channel is selected for masking at a time, and the nuclear norm of all the channels, along with the masked channel, is re-calculated. The CI score of the selected channel \(c\) is computed as the difference between the two nuclear norms, as outlined in equation (\ref{eq:8}), where \(f^{(l)}\) represents the feature maps of convolutional layer \(l\), \(f_{c}^{(l)}\) denotes the feature map of channel \(c\) in convolutional layer \(l\), and \(M_{c}\) is a masked matrix of similar size as \(f_{c}^{(l)}\), containing ones in all rows except the \(c^{th}\) row, which is filled with zeros. The masking of the \(c^{th}\) row in \(f_{c}^{(l)}\) is carried out by element-wise multiplication of \(f_{c}^{(l)}\) with \(M_c\). However, to obtain the final score, the CI scores of all selected channels are averaged across all batches in the training dataset.
\begin{equation} \label{eq:8}
    CI(c) = \|f^{(l)}\|_* - \|f_{c}^{(l)} \odot M_{c}\|_*
\end{equation}
The CI scores for all the channels in all the convolutional layers of the backbone network and the PrimaryCaps layer are computed. The lower CI score of any channel in the layer signifies the least independent channel and contains less information, so it can be pruned without impacting the network. Hence, the CI scores for all the channels of convolutional layers are then sorted in ascending order, and the channels with the lowest CI scores are selected for pruning from the ordered list of batch-averaged CI scores.
% \begin{figure}[htbp!]
%     \centering
%     \includegraphics[width=0.7\linewidth]{Channel_Pruning_With_Chip.png}
%     \caption{\label{fig:2}Channel Pruning with CHIP}
% \end{figure}

The number of channels to be pruned is decided using a pruning ratio. However, to ensure enough channels are available after pruning to rebuild the PrimaryCaps layer with a pre-defined capsule dimension D, the channels in all the convolutional layers and the PrimaryCaps layer are pruned in multiple of D. For example, 
%in Fig. \ref{fig:2}, 
if the number of channels is 12 and the pruning ratio is 0.25, ideally, three channels should be pruned. However, if the number for capsule dimensions is four, after pruning three channels, only nine channels are left out, and since it is not a multiple of four, instead of pruning three channels, four channels with the lowest CI scores are pruned to have eight output channels, which is a multiple of D. Additionally, pruning is skipped for a convolutional layer with number of channels less than D. This is critical for rebuilding the model - after pruning - with the pruned backbone network and the pruned PrimaryCaps.

\subsection{The Interplay between Vanishing Activation and Pruning }
We built and trained a base CapsNet (BaseCapsNet) model with a backbone network consisting of three convolutional layers, a PrimaryCaps layer, and a ClassCaps layer over multiple epochs using the total margin loss given by equation (\ref{eq:6}). The model with the highest accuracy is selected and then undergoes re-training with the total loss function defined in equation (\ref{eq:5}), which includes both margin and CCM losses because CCM supports structure pruning. As illustrated in equation (\ref{eq:8}), CI scores are calculated for all the convolutional channels in the network using the CHIP algorithm, and the channels with the lowest CI scores are then identified to be pruned.

After setting the pruning ratio, the lowest-scored channels of the convolutional layers in both the backbone network and the PrimaryCaps layer of the best pre-trained model are pruned. Then, the model is rebuilt using the remaining channels of the backbone network and the primary capsules with a new ClassCaps layer, which ensures the rerouting of the pruned PrimaryCaps to appropriate ClassCaps. The pruned CapsNet model is then retrained - over multiple epochs - using only the total margin loss given by equation (\ref{eq:6}). The model with the highest test accuracy, named the best-pruned model, is saved for later comparison of the performance of different pruning ratios. The process of setting the pruning ratio, carrying out model pruning, followed by model rebuilding, model training, and validation, is repeated multiple times for different pruning ratios of choice.

To recreate the vanishing activations issue typically seen in the CapsNet \cite{Everett2023}, IntermediateCaps layers are added one by one between the PrimaryCaps and the ClassCaps layers. At first, one IntermediateCaps layer is inserted, and the model is trained using the total margin loss. After training, if the validation accuracy is quite reasonable and changes considerably over the epochs, an additional IntermediateCaps layer is inserted, and the new model is trained. During model training, the outputs of the capsules in an IntermediateCaps layer are averaged over all the input images, and then the Frobenius norm is calculated along the capsule dimension to obtain individual capsule activations. When the activation value is less than or equal to 0.01 for a capsule, it is pronounced dead.

The IntermediateCaps layers are continuously added until the model accuracy reduces to a meagre value ($\sim10\%$) over multiple epochs. When a very low non-recovering validation accuracy is obtained, the model is assumed to be broken due to infinitesimal capsule activations in the IntermediateCaps layers. 
% This assumption is confirmed by plotting the activation of the capsules in all the IntermediateCaps layers. 
Once a model breaks due to vanishing activations, the backbone network and the PrimaryCaps layer are replaced with the best-pruned model's backbone network, and PrimaryCaps are obtained with the smallest pruning ratio, and the model is re-trained. Suppose the pruned model continues to suffer from vanishing activations. 
% Then, the pruned backbone network and the pruned PrimaryCaps obtained with larger pruning ratios are used subsequently.
Then, the model is rebuilt and trained with a larger pruning ratio.

In this work, all CapsNet models are trained and tested using the well-known CIFAR10 dataset \cite{Krizhevsky09learningmultiple}, which has 60,000 coloured images of size 32×32 pertaining to 10 different classes. Out of the total images in the dataset, 50,000 are used for training, and 10,000 are used for testing. The Capsule Network models are built with the PyTorch library \cite{Paszke2019}. For building the Capsule Networks and carrying out model training and evaluation, Google Colab is used with the GPUs and TPUs. 

\section{Experimental Results and Discussions}
% \subsection{Pruning Performance Analysis}
Six different base model configurations without any IntermediateCaps layers are pruned with seven different pruning ratios (PR) to analyze the effect of pruning on model performance. 
%Their accuracies are plotted in Fig. \ref{fig:3}. 
The model accuracies obtained for different pruned model configurations are summarized in Table \ref{tab:1}. Three important inferences can be made from the table. Firstly, for all six model configurations, when the pruning ratio is increased from 0.125 to 0.75, the model accuracy increases at the beginning, reaches a peak and then starts plummeting. For instance, the unpruned [C=10, D=8] model accuracy is 73.92\% (C is the number of capsules and D is the dimension of the capsules). As the pruning ratio increases, the model accuracy also increases and reaches a peak accuracy of 74.71\% at a pruning ratio PR=0.5. Then, the accuracy falls to 73.84\% (which is below the unpruned model accuracy) when the PR is 0.75. Thus, the pruning ratio must be chosen wisely to achieve the best model performance.

\begin{table}
\centering
\caption{\label{tab:1}BaseCapsNet models accuracies (in \%) achieved with different pruning ratios}
\resizebox{\linewidth}{!}{%
\begingroup
\setlength{\tabcolsep}{5pt}
\begin{tabular}{|c|ccccccc|}
\hline
Model Configuration & PR=0.0 & PR=0.125 & PR=0.25 & PR=0.375 & PR=0.5 & PR=0.60/0.625* & PR=0.75 \\
\hline
{[}C=10, D=8{]}     & 73.92  & 74.26    & 74.37   & 74.53    & 74.71  & 74.09          & 73.84   \\
{[}C=20, D=8{]}     & 77.05  & 78.71    & 77.75   & 77.61    & 76.76  & 76.88          & 76.12   \\
{[}C=32, D=8{]}     & 77.37  & 78.01    & 76.88   & 78.50    & 78.50  & 77.24          & 76.63   \\
{[}C=16, D=6{]}     & 75.40  & 75.54    & 75.39   & 74.66    & 73.99  & 73.02*         & 71.49   \\
{[}C=16, D=12{]}    & 76.30  & 79.12    & 78.43   & 78.43    & 78.21  & 77.02*         & 75.06   \\
{[}C=8, D=12{]}     & 75.12  & 76.12    & 75.46   & 75.44    & 74.94  & 75.39*         & 74.28  \\
\hline
\end{tabular}
\endgroup
}
\end{table}
% \begin{figure}[htbp!]
%     \centering
% \begin{multicols}{2}
%     \includegraphics[width=.85\linewidth]{Plots/C10D8.png}\par 
%     \includegraphics[width=.85\linewidth]{Plots/C20D8.png}\par \end{multicols}\begin{multicols}{2}
%     \includegraphics[width=.85\linewidth]{Plots/C32D8.png}\par
%     \includegraphics[width=.85\linewidth]{Plots/C16D6.png}\par \end{multicols} \begin{multicols}{2}
%     \includegraphics[width=.85\linewidth]{Plots/C16D12.png}\par 
%     \includegraphics[width=.85\linewidth]{Plots/C8D12.png}\par
% \end{multicols}
% \caption{\label{fig:3}Accuracies of unpruned and pruned BaseCapsNet models}
% \end{figure}
Secondly, keeping the capsule dimensions D fixed when the number of capsules C is doubled, the model accuracy always increases irrespective of the pruning ratio set. It is evident that when D is 8 and C is changed from 10 to 20, the model accuracy goes up from 74.26\% to 78.71\% at PR=0.125. Similarly, when D is 12 and C is doubled from 8 to 16, the model accuracy improves from 75.44\% to 78.43\% at PR=0.375. 

The third and last point is when the number of capsules C is fixed, and the number of capsule dimensions is doubled, the model accuracy increases despite the pruning ratio. This means that the more dimensions the capsules in a CapsNet have, the better the model performance will be. This is concluded by comparing the model accuracies of the configurations [C=16, D=6] and [C=16, D=12] for different pruning ratios. As we see, when PR=0.125, the model with a D of 6 gets an accuracy of 75.54\%, whereas the model with a D of 12 achieves a higher accuracy of 79.12\%.
\subsection{Effect of Intermediate Capsule Layers}
Up to ten IntermediateCaps layers (I) are added for four different model configurations to assess the advantages of going deeper with CapsNet. Adding IntermediateCaps layers to CapsNet models increases the model accuracy for most of the models up to five IntermediateCaps layers, which is evident in Table \ref{tab:2}. On another note, after adding seven IntermediateCaps layers to CapsNet[C=20, D=8], the validation accuracy of the model drastically drops to 10\% showing the model’s incapability to learn further. Thus, the CapsNet[C=20, D=8] model is selected for the in-depth evaluation of the vanishing activations.
\begin{table}
\centering
\caption{\label{tab:2}Comparison of the accuracies (in \%) of CapsNet models with an increasing number of intermediate (I) capsule layers}
\resizebox{\linewidth}{!}{
\begingroup
\setlength{\tabcolsep}{5pt}
\begin{tabular}{|c|ccccccccccc|}
\hline
\textbf{Model Configuration} & \textbf{I=0}& \textbf{I=1} & \textbf{I=2} & \textbf{I=3} & \textbf{I=4} & \textbf{I=5} & \textbf{I=6} & \textbf{I=7} & \textbf{I=8} & \textbf{I=9} & \textbf{I=10} \\
  \hline
{[}C=10, D=8{]}  & 73.9 & 75.2 & 75.2 & 75.5 & 75.4 & 74.3 & 74.1 & 75.2 & 74.4 & 73.3 & 73.5 \\
{[}C=20, D=8{]} & 77.1 &76.7 &77.3 &77.7 &77.4 &76.8 &76.2 &{\color[HTML]{FF0000} 10} &
  {\color[HTML]{FF0000} 10} &
  {\color[HTML]{FF0000} 10} &
  {\color[HTML]{FF0000} 10} \\
{[}C=16, D=12{]} & 76.3 & 78.6 & 78.6 & 78.2 & 77.1 & 77.6 & 76.7 & 76.4 & 75   & 75.6 & 74.3 \\
{[}C=8, D=12{]}  & 75.1 & 75.1 & 75.2 & 75.4 & 75.1 & 75.5 & 75.1 & 75.3 & 73.7 & 75.3 & 75.2 \\
\hline
\end{tabular}
\endgroup
}
\end{table}
\subsection{Exploring Vanishing Activations}
\begin{figure}
    \centering
\begin{multicols}{2}
\includegraphics[width=.4\textwidth, height=2.2cm]{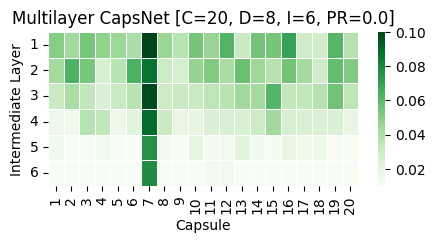} \subcaption{}
\includegraphics[width=.4\textwidth, height=2.2cm]{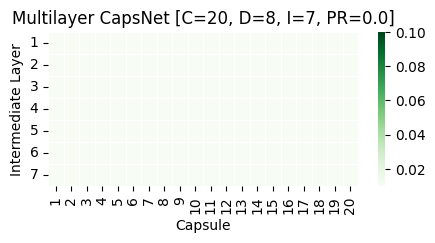}\subcaption{}
\end{multicols}
\caption{\label{fig:4}Activations heatmap of the CapsNet[C=20, D=8] when trained with intermediate capsule layers (a) I=6 and (b) I=7}
\end{figure}

\begin{figure}
    \centering
    \includegraphics[width=.4\textwidth, height=2.8cm]{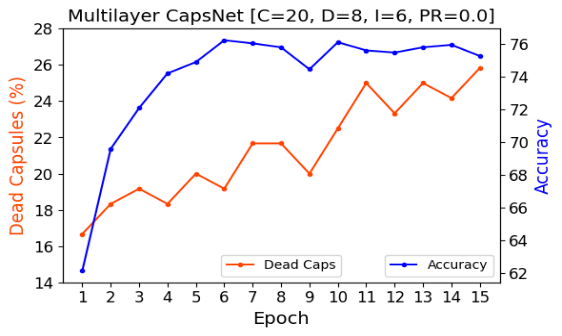}
    \caption{\label{fig:5}The trend for dead capsules and model accuracy seen in CapsNet[C=20, D=8, I=6]}
\end{figure}

To view the vanishing activations in the CapsNet[C=20, D=8] models with 6 and 7 IntermediateCaps layers, heatmaps are plotted for the capsule activations of the IntermediateCaps layers as shown in Fig. \ref{fig:4}. Unlike the case of the CapsNet[C=20, D=8] with I=6 IntermediateCaps layers, where most of the capsule activations in layers 5 and 6 are close to 0.01, for the CapsNet[C=20, D=8] with I=7 IntermediateCaps layers, all the capsule activations in all the IntermediateCaps layers are found to be zero. When the model breaks in the case of I=7, all the capsules in the IntermediateCaps layers are dead, resulting in a poor validation accuracy of 10\%. Increasing the number of intermediate capsule layers beyond seven does not help recover the dead capsules, and hence, validation accuracy remains at 10\% thereafter. In addition, Fig. \ref{fig:5} shows how the percentage of dead capsules increases for CapsNet[C=20, D=8, I=6] over the training epochs while validation accuracy mostly stays around 76\% beyond epoch 6.

\subsection{Going deeper with pruning}
From Table \ref{tab:3}, it can be inferred that the validation accuracies of the BaseCapsNet[C=20, D=8] pruned with pruning ratios (PR) of 0.125, 0.25 and 0.375 are better than the validation accuracy achieved without pruning. Hence, 0.125, 0.25 and 0.375 pruning ratios are only used to overcome the vanishing activations problem. 
\begin{table}
\centering
\caption{\label{tab:3}Comparison of pruned Multilayer CapsNet{[}C=20, D=8{]} accuracies (in \%) for Intermediate (I) capsule layers}
\begingroup
\setlength{\tabcolsep}{6pt}
\begin{tabular}{|c|cc|}
\hline
\textbf{Pruning Ratio} & \textbf{I=6} & \textbf{I=7}              \\
\hline
0                      & 76.2         & {\color[HTML]{FF0000} 10} \\
0.125                  & 78.9         & 79.3                      \\
0.25                   & 79.5         & 79.6                      \\
0.375                  & 79           & 79.2                       \\
\hline
\end{tabular}
\endgroup
\end{table}

\begin{figure}
    \centering
    \begin{multicols}{3}
    \includegraphics[width=.33\textwidth]{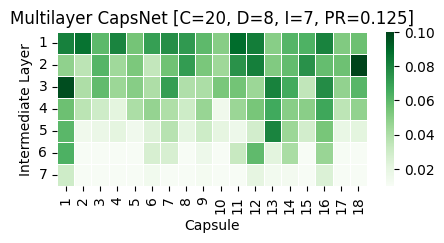}\subcaption{}

    \includegraphics[width=.33\textwidth]{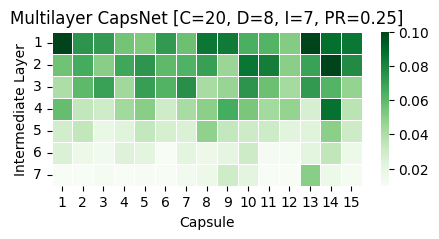}\subcaption{}
    
    \includegraphics[width=.33\textwidth]{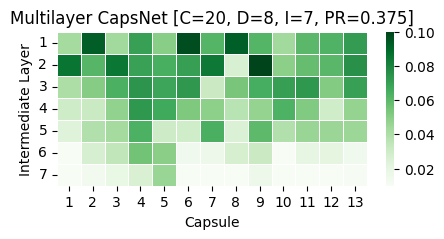}\subcaption{}
    \end{multicols}
    \caption{\label{fig:6}Activation heatmaps for CapsNet[C=20, D=8, I=7] pruned with (a) 0.125, (b) 0.25 and (c) 0.375 ratios}
\end{figure}
\begin{figure}
    \centering
    \begin{multicols}{2}
    \includegraphics[width=.42\textwidth, height=3cm]{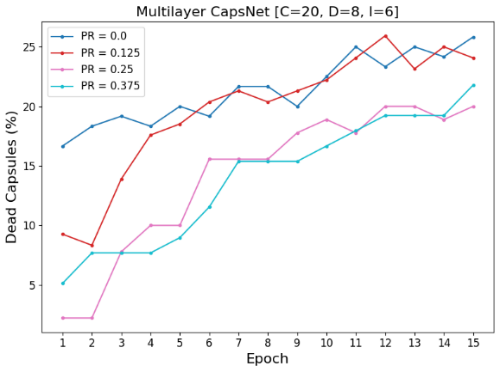}\subcaption{}
    
    \includegraphics[width=.42\textwidth, height=3cm]{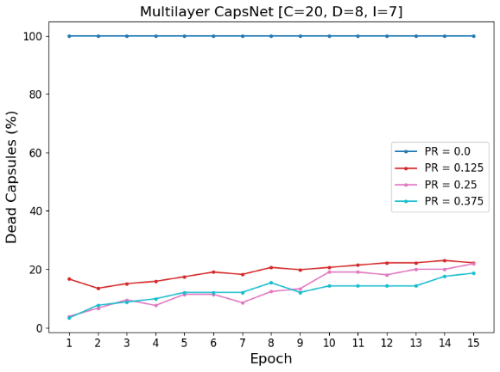}\subcaption{}
        \end{multicols}
    \caption{\label{fig:7}Comparison of percentages of dead capsules present in CapsNet[C=20, D=8] with different pruning ratios for (a) 6 intermediate capsule layers and (b) 7 intermediate capsule layers}
\end{figure}

For the CapsNet[C=20, D=8, I=7] model, as the pruning ratio increases, more capsules in its IntermediateCaps layers 6 and 7 become active, which is obvious from the activation heatmaps shown in Fig. \ref{fig:6}. In other words, pruning reduces the percentage of dead capsules in the models CapsNet[C=20, D=8, I=6] and CapsNet[C=20, D=8, I=7], as shown in Fig. \ref{fig:7}. Also, it is noted from Table \ref{tab:3} that with all three pruning ratios, the validation accuracies of the CapsNet[C=20, D=8, I=6] model increased by 3.3\%. For the CapsNet[C=20, D=8, I=7] model, there is a drastic increase in the model accuracy from 10\% to 79\%, which shows how well pruning can mitigate vanishing activations.
\begin{figure}
    \centering
    \includegraphics[width=.55\linewidth]{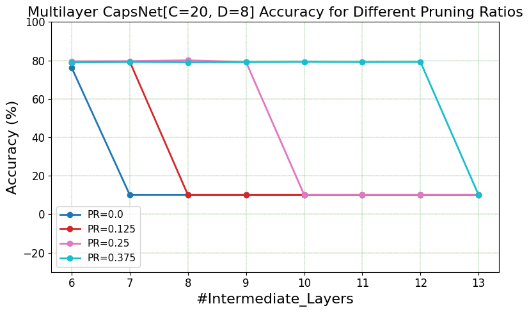}
    \caption{\label{fig:6}Accuracy of Multilayer CapsNet[C=20, D=8] for different pruning ratios }
\end{figure}
There is no doubt that model pruning alleviates the vanishing activations. However, lower pruning ratios are not largely helpful while building deeper CapsNet. From Fig. \ref{fig:6}, it can be observed that without pruning, a maximum of six intermediate layers could be added, and with pruning ratios of 0.125 and 0.25, a maximum of seven and nine intermediate layers could be supported. And 12 intermediate layers can be added when the pruning ratio is 0.375. Thus, the higher the pruning ratio is, the more intermediate layers we can add to the CapsNet while building deeper CapsNet.

\subsection{Other Experiments}
\subsubsection{\textbf{Different Kernel Sizes.}} Table \ref{tab:5} presents the accuracies obtained by the BaseCapsNet built with different convolutional layers for the backbone network and the PrimaryCaps. The data in Table \ref{tab:5} shows that maximum accuracy is achieved for kernel size 5 and 32 8D capsules. However, training the [C=32, D=8]model with a kernel size of 5 and multiple intermediate layers became computationally expensive due to resource constraints. As a result, models with at most 20 capsules and a kernel size of 3 are selected for this project.
\begin{table}
\centering
\caption{\label{tab:5}Accuracy of BaseCapsNet trained with different convolutional configurations}
\resizebox{\linewidth}{!}{
\begingroup
\begin{tabular}{|c|c|c|c|c|c|c|c|c|}
\hline
\multicolumn{1}{|c|}{\multirow{2}{*}{\textbf{Model}}} &
  \multicolumn{4}{c|}{\textbf{Backbone Network}} &
  \multicolumn{3}{c|}{\textbf{PrimaryCaps}} &
  \multicolumn{1}{c|}{\multirow{2}{*}{\textbf{Accuracy (\%)}}} \\
\multicolumn{1}{|c|}{} &
  \multicolumn{1}{c}{\textbf{Layers}} &
  \multicolumn{1}{c}{\textbf{Kernel}} &
  \multicolumn{1}{c}{\textbf{Stride}} &
  \multicolumn{1}{c|}{\textbf{Padding}} &
  \multicolumn{1}{c}{\textbf{Kernel}} &
  \multicolumn{1}{c}{\textbf{Stride}} &
  \multicolumn{1}{c|}{\textbf{Padding}} &
  \multicolumn{1}{c|}{} \\
  \hline
{[}C=32, D=8{]} & 3$\times$(Conv2d+Batch Normalization+ReLU)& 5,5,5 & 1,1,1 & 0,0,0 & 5 & 1 & 0 & 82.17 \\
{[}C=32, D=8{]} & 3$\times$(Conv2d+Batch Normalization+ReLU)& 3,3,3 & 1,2,1 & 1,1,1 & 3 & 2 & 0 & 77.37 \\
{[}C=20, D=8{]} & 3$\times$(Conv2d+Batch Normalization+ReLU)& 5,5,5 & 1,1,1 & 0,0,0 & 5 & 1 & 0 & 72.94 \\
{[}C=20, D=8{]} & 3$\times$(Conv2d+Batch Normalization+ReLU)& 3,3,3 & 1,2,1 & 1,1,1 & 3 & 2 & 0 & 77.05 \\
\hline
\end{tabular}
\endgroup
}
\end{table}
% \subsubsection{\textbf{Different Alpha for CCM Loss.}} The BaseCapsNet models are trained with CCM loss before pruning. To control the contribution of the CCM loss towards the total model loss, a constant, alpha is used as shown in equation (\ref{eq:6}). The CapsNet[C=16, D=12] model is trained with five different values of alpha, i.e., 0.1, 0.01, 0.05, 0.001, 0.005, and the validation accuracies and losses obtained with them are compared in Fig. \ref{fig:9}. Because the highest accuracy and lowest loss are achieved with the alpha 0.01, it is selected for training all other models with CCM loss.
% \begin{figure}
%     \centering
%     \includegraphics[width=0.45\textwidth, height=2.8cm]{Plots/C16D12_Acc_Los_CCM_Alpha.png}
%     \caption{\label{fig:9}CapsNet[C=16, D=12] mean accuracy and loss for different CCM alpha values}
% \end{figure}
\subsubsection{\textbf{Maximum Intermediate Layers.}} CapsNet models with different numbers of capsules and capsule dimensions are built to determine how many intermediate layers can be supported between the PrimaryCaps and ClassCaps layers. An important observation can be made from Table \ref{tab:4}, i.e., reducing the capsule dimensions while keeping the number of capsules constant and reducing the number of capsules while keeping the capsule dimensions constant results in adding a lesser number of intermediate layers to the CapsNet. Thus, a possible option to go deeper would be to increase the capsule dimensions while keeping the number of capsules constant in the CapsNet.
\begin{table}
\centering
\caption{\label{tab:4}Maximum depth for different CapsNet configurations possible without pruning}
\resizebox{\linewidth}{!}{
\begingroup
\begin{tabular}{|>{\centering\arraybackslash}p{1in}|>{\centering\arraybackslash}p{2in}|>{\centering\arraybackslash}p{2.5in}|} 
\hline
\textbf{Capsules {[}C{]}} & \textbf{Capsule Dimensions {[}D{]}}& \textbf{Maximum Intermediate Layers {[}I{]}}\\ 
\hline
20 & 10 & 8                \\ 
20 & 8  & 6                \\ 
20 & 6  & 4                \\ 
16 & 6  & 5                \\ 
10 & 6  & 7                \\ 
10 & 8  & \textgreater{}30          \\ 
\hline
\end{tabular}
\endgroup
}
\end{table}
\section{Conclusion and Future Work}
In this paper, Capsule Networks with varying numbers of capsules and capsule dimensions are pruned using six different pruning ratios. Channel pruning not only reduced the number of model parameters and computational complexity but also significantly increased the model accuracy. 
% Doubling the number of capsules while keeping the capsule dimensions constant or doubling the capsule dimensions while keeping the number of capsules fixed results in around a 5\% improvement in the model accuracy, irrespective of the pruning ratio used. 
When up to 10 intermediate layers are added to four different unpruned models to assess the benefits of building deeper Capsule Networks, a slight improvement in the model accuracies is observed. %However, one of the four models showed a sudden drop in accuracy to around 10\% after adding 7 intermediate layers. %Upon comparing the capsule activations of the same model for 6 and 7 intermediate layers, it is discovered that all the intermediate capsules in the 7 intermediate layer model are dead.
%This led to the research for other model configurations, which also showed decreased accuracy after adding multiple intermediate capsule layers. 
For a CapsNet with a large number of capsules, it is found that when the number of capsule dimensions is reduced, most of the capsule activations shrink to zero. For example, the model with 20 capsules of 8 dimensions breaks down after adding 7 intermediate layers, whereas the model with 10 capsules of 8 dimensions does not break even after adding 30 intermediate layers.
Finally, in this paper, CCM loss-based channel pruning is used to mitigate the vanishing capsule activations in deeper Capsule Networks, revealing that a pruned model has fewer dead capsules than a deep unpruned model. In conclusion, increasing the pruning ratio allows for building deeper Capsule Networks without compromising model performance.
% \section{Future Work}
% In this project, bigger Capsule Networks with more than 20 capsules could not be explored due to a lack of computational resources. 
% In future, bigger models can be built to generalize the contribution of channel pruning toward mitigating the vanishing activations, which is already shown for relatively small Capsule Networks in this paper.
% The backbone network built in this project is not a state-of-the-art network like ResNet-50 or VGG-16, and only the CIFAR10 dataset is used for model training and validation. Thus, this paper lacks benchmarking. Hence, efforts can be made to build Capsule Networks with a backbone comprising state-of-the-art models and train them with widely known datasets like ImageNet, TinyImageNet, and AFFNIST.
In the future, bigger models can be built to generalize channel pruning's contribution to mitigating vanishing activations, as this paper already shows for relatively small Capsule Networks. Additionally, it would be exciting to see how other pruning techniques mitigate vanishing activations in deeper Capsule Networks. The source code for this project is available on the GitHub \href{https://github.com/sidsahu88/leeds_msc_ocom5300m_project}{link}.
\bibliography{main}
\end{document}